\documentclass[runningheads]{llncs}
\usepackage{graphicx}

\usepackage{tikz}
\usepackage{comment}
\usepackage{amsmath,amssymb} 
\usepackage{color}

\usepackage[accsupp]{axessibility}  

\usepackage{booktabs}       
\usepackage{amsfonts}       
\usepackage{nicefrac}       
\usepackage{microtype}      
\usepackage{amsfonts}       
\usepackage{float}
\usepackage{sidecap}
\usepackage{tikz}
\usepackage{amsmath}
\usepackage{wrapfig}
\definecolor{colour}{RGB}{228, 24, 133}

\begin{document}
\pagestyle{headings}
\mainmatter
\def\ECCVSubNumber{4195}  

\title{Word-Level Fine-Grained Story Visualization} 

\titlerunning{Word-Level Fine-Grained Story Visualization}
%
\author{Bowen Li\inst{1}\index{Li, Bowen} \and Thomas Lukasiewicz\inst{2,1}\index{Lukasiewicz, Thomas}}
\authorrunning{L. Bowen and L. Thomas}
%
\institute{University of Oxford, UK \\
\email{firstname.lastname@cs.ox.ac.uk}
\and 
Institute of Logic and Computation, TU Wien, Austria}

\maketitle

\begin{abstract}
Story visualization aims to generate a sequence of images to narrate each sentence in a multi-sentence story with a global consistency across dynamic scenes and characters. Current works still struggle with output images' quality and consistency, and rely on additional semantic information or auxiliary captioning networks. To address these challenges, we first introduce a new sentence representation, which incorporates word information from all story sentences to mitigate the inconsistency problem. Then, we propose a new discriminator with fusion features and further extend the spatial attention to improve image quality and story consistency. Extensive experiments on different datasets and human evaluation demonstrate the superior performance of our approach, compared to state-of-the-art methods, neither using segmentation masks nor auxiliary captioning networks.
\end{abstract}

\section{Introduction}
Image generation from different-modal text descriptions with semantic alignment is a challenging task and has the potential for many applications, including art creation, computer-aided design, and image editing. Recently, due to the great progress in realistic image generation based on generative adversarial networks (GANs)~\cite{goodfellow2014generative}, text-guided image generation and modification have drawn much attention~\cite{reed2016generative,zhang2017stackgan,xu2018attngan,li2019controllable,zhu2019dm,hinz2019semantic,li2020image,li2020manigan,li2020lightweight,nam2018text,dong2017semantic}. 

\begin{figure}[t]
\centering
\begin{minipage}{0.98\textwidth}
\includegraphics[width=1\linewidth, height=0.736\linewidth]{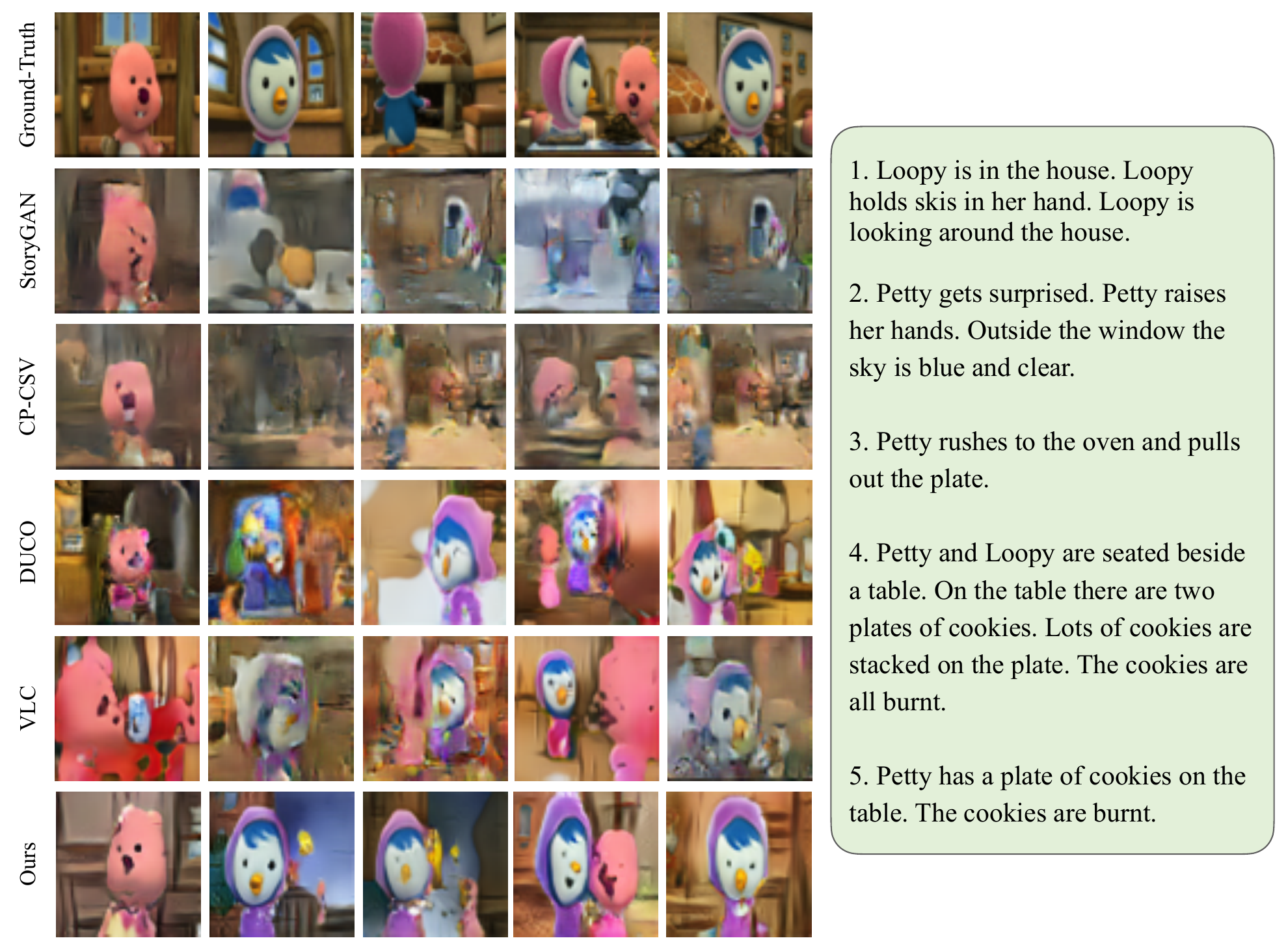}
\end{minipage}

\centering
\vspace{-2ex}
\caption{Examples of story visualization on different methods, with the given story sentences and ground-truth story images.}
\label{fig:example}
\vspace{-3ex}
\end{figure}

Differently from text-to-image generation, story visualization aims to generate a sequence of story images given a multi-sentence story, which is more challenging, as it further requires output images to be consistent, e.g., having a similar background or objects' appearances. StoryGAN~\cite{li2019storygan} first proposed a sequential conditional GAN-based framework. Using StoryGAN as the backbone, CP-CSV~\cite{song2020character} utilized segmentation masks to keep character consistency, and DUCO~\cite{maharana2021improving} and VLC~\cite{maharana2021integrating} adopted additional auxiliary captioning networks to improve text-image semantic alignment. However, all these works still struggle with the quality of output images, and may fail to generate fine-grained image regional details corresponding to different-semantic words (see Fig.~\ref{fig:example}). 
Besides, CP-CSV requires character segmentation masks in the network, which are hard to get, and the performance of DUCO and VLC may be affected by auxiliary captioning networks. Moreover, some keywords, e.g., describing the global style of a story, do not exist in all sentences within a story. So, if keywords only appear in some sentences, these sequential generation methods may fail to keep consistent object appearances and a global style in all synthetic images. 

To tackle these problems, we first introduce a new sentence representation, where each sentence representation in a story can selectively incorporate different word information from the entire story to ensure a global consistency. 
Then, we propose a new discriminator with fusion features, which provides the generator with fine-grained training feedback, evaluating the quality of fusion features to examine whether all word-required attributes are generated in the story images. Finally, we further explore word-level spatial attention~\cite{xu2018attngan} in story visualization, which can not only highlight local word-related image regions in the generation process, but also capture long-distant correlations between regions in the current image and words from other sentences within the same story. By doing this, the generator can focus on different parts of an image to enable better regional details and also to ensure consistency over all synthetic images in a story. The main contributions are briefly summarized as follows:

\noindent $\bullet$  We propose a new approach for story visualization, which contains three novel components: new sentence representation, discriminator with fusion features, and extended spatial attention, neither utilizing cost segmentation masks nor additional auxiliary captioning networks. 

\noindent $\bullet$ Our approach builds a new state of the art on Pororo-SV, measured by different evaluation metrics with large absolute margins, and further establishes a strong benchmark on Abstract Scenes.

\noindent $\bullet$ We conduct extensive experiments, including a human evaluation, to demonstrate the superior performance of our approach over baselines, in terms of image quality, text-image semantic alignment, and story consistency. The code is available at \textcolor{colour}{https://github.com/mrlibw/Word-Level-Story-Visualization}.

\section{Related Work}
Story visualization aims to generate a sequences of images corresponding to a multi-sentence story, and imposes consistency over output story images. StoryGAN~\cite{li2019storygan} first proposed this task and also introduced a GAN-based sequential generation network. CP-CSV~\cite{song2020character} was built on StoryGAN, and utilized segmentation masks to improve the character consistency. DUCO~\cite{maharana2021improving} and VLC~\cite{maharana2021integrating} also adopted StoryGAN as the backbone, and added auxiliary captioning networks to build a text-image-text circle. However, all these methods still struggle with the quality of output images, and fail to generate fine-grained regional details.
Our approach explores word-level information within the story, to ensure a high-quality image generation with good text-image semantic alignment and story consistency, neither requiring an additional supervision from segmentation masks, like CP-CSV, nor auxiliary captioning networks, like DUCO and VLC.

Text-to-image generation is closely related to our work, which generates one image from one given text description and keeps semantic alignment between image and text~\cite{reed2016generative,zhang2018stackgan++,qiao2019mirrorgan,hinz2019semantic,zhu2019dm,xu2018attngan,li2021memory,tao2020df}.  
AttnGAN~\cite{xu2018attngan} and ControlGAN\cite{li2019controllable} introduced word-level attention to fuse text and image features. DMGAN\cite{zhu2019dm} proposed a dynamic memory module to refine image features. DF-GAN\cite{tao2020df} introduced a deep text-image fusion block to fuse text and image information. OPGAN~\cite{hinz2019semantic} relied on additional semantic information to improve output results. XMC-GAN~\cite{zhang2021cross} is a contrastive-learning-based method with a single stage. DALL-E~\cite{ramesh2021zero} proposed a transformer-based method for zero-shot text-to-image generation. Li et al.~\cite{li2021memory} proposed a semi-parametric approach via constructing a memory bank of image features, selectively fusing stored image features into the generation pipeline.

Another related work is video generation from text~\cite{li2018video,pan2017create,gupta2018imagine,balaji2019conditional,mahon2020knowledge}, which generates continuous frames from only one text input. Differently, story visualization does not require the frames to flow continuously, and allows synthetic story images to be discrete with different scene views. 

\section{Fine-Grained Story Visualization}
The model aims to produce a series of story images from a given multi-sentence story, one for each, and output images should be realistic, semantically match corresponding sentences, and keep consistency. To achieve this, we propose three novel components: (1) new sentence representation that selectively fuses word information from an entire story, (2) discriminator with fusion features, and (3)~extended spatial attention at story level. 
\begin{figure}[t]

\begin{minipage}{1\textwidth}
\includegraphics[width=1\linewidth, height=0.1513\linewidth]{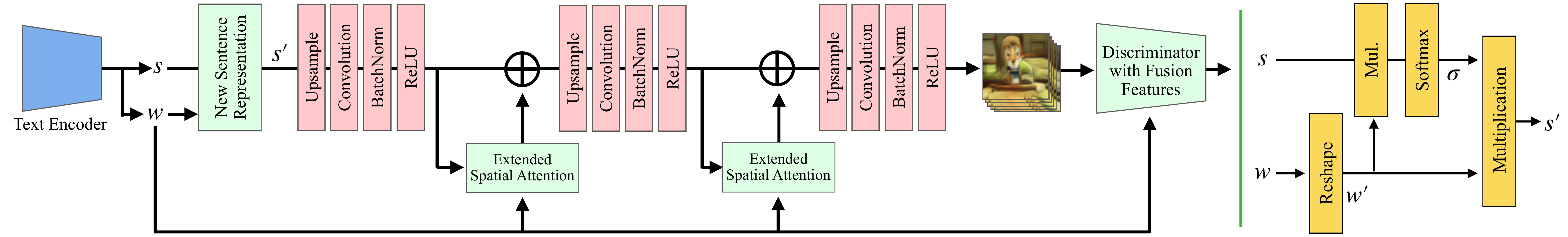}
\end{minipage}

\centering
\vspace{-2ex}
\caption{Left: architecture of the proposed approach. Right: design of the proposed new sentence representation.}
\label{fig:archi}
\end{figure}
\subsection{Architecture}
Similarly, we adopt StoryGAN~\cite{li2019storygan} as our basic backbone. To produce story images with fine-grained regional details and a better consistency, we propose to utilize both global sentence vectors and fine-grained word embeddings in the generation pipeline. 
However, there is only one discriminator, and neither segmentation masks nor auxiliary captioning networks are used in our method.

Given a multi-sentence story $Z$ with a sequence of $n$ story descriptions, $S_1,\ldots,S_n$, a text encoder, e.g., bi-directional LSTM~\cite{xu2018attngan}, encodes the story sentences into a sequence of sentence vectors $s\in \mathbb{R}^{N \times D} $ with corresponding word embeddings $w \in \mathbb{R}^{N \times D \times L}$, where $N$ is the number of sentences in a story, $D$ is the feature dimension, and $L$ is the number of words in a sentence. 
Then, we feed both sentence vectors and word embeddings into the generation pipeline using a series of upsampling blocks to produce story images at the required resolution. 
To generate high-quality images with fine-grained regional details, we propose to utilize word-level information in the network, including the input representation, the generator, and the discriminator.

\subsection{Sentence Representation with Word Information}
One problem arising in previous works~\cite{li2019storygan,song2020character} is that they feed given story sentences sequentially into the generation pipeline to produce corresponding story images. Thus, when some keywords do not appear in all sentences, the corresponding keyword-related attributes may not be generated in all story images due to the sequential generation, and then the synthetic story may fail to keep consistency across all images. 

To address the issue, we propose a new sentence representation, which can selectively incorporate different word information from the entire story to mitigate the inconsistency problem. To build this sentence representation, we first reshape the word embeddings $w \in \mathbb{R}^{N \times D \times L}$ to get $w' \in \mathbb{R}^{D \times (L\ast N)}$. Then, we calculate the correlation weights $\sigma\in \mathbb{R}^{N\times (L\ast N)}$ between each sentence and all words within the story, according to applying a matrix multiplication followed by a Softmax operation between $w'$ and a sentence vector $s \in \mathbb{R}^{N \times D}$, denoted as $\sigma = \text{Softmax}(sw')$. Then, a new sentence representation weighted by word-sentence relation can be obtained by doing $s' = \sigma(w')^T$, and is fed into the generation pipeline at the beginning. By doing this, this new sentence vectors can selectively incorporate different word information, even if they only appear in some sentences, to keep information consistency across all story images.

\subsection{Discriminator with Fusion Features}
\label{sec:dis}
\begin{figure}[h]
\vspace{-3ex}
\centering
\begin{minipage}{1\textwidth}
\includegraphics[width=1\linewidth, height=0.3125\linewidth]{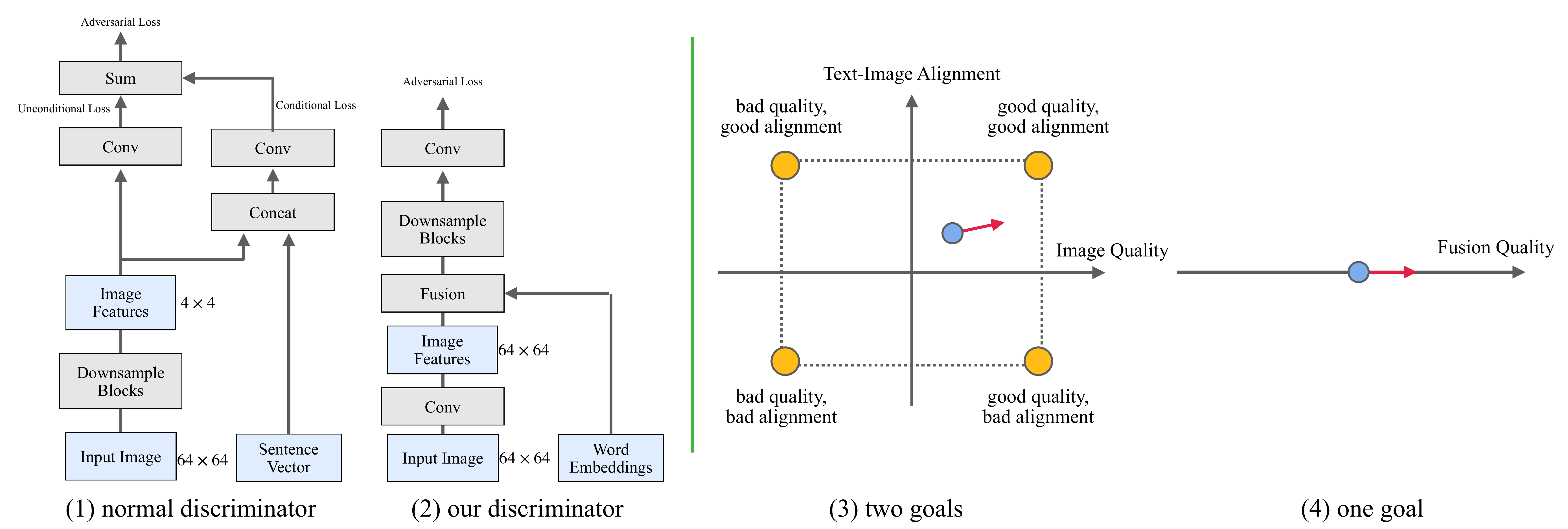}
\end{minipage}

\centering
\vspace{-2ex}
\caption{Left: comparison between the normal discriminator (1) and ours (2). Right: a diagram for two goals of current methods (3) and the goal of our proposed one-way output design (4).}
\label{fig:archi_dis}
\vspace{-3ex}
\end{figure}
In this work, we propose a new discriminator with one-way output, which works on fusion features that contain both image and text information. Although DFGAN~\cite{tao2020df} also introduced a one-way output discriminator, differently, (1) our discriminator uses fine-grained word-level information and image features at the original size without information loss, while DFGAN first downsamples image features to a small size at $4 \times 4$, which may suffer a potential information loss problem, (2) our discriminator builds fusion features via applying matrix multiplication between detailed image features and word-level text representations, while DFGAN concatenates a coarse sentence vector with small-scale image features at $4 \times 4$, and may fail to comprehensively explore the correlation between image and text representations, and (3) loss objectives for our discriminator only check the difference of fusion features created from word representations and synthetic images and fusion features created from word representations and real images, while fusion features in DFGAN promote an additional gradient penalty, along with discriminator objectives.

\subsubsection{Fusion Features.}
As shown in Fig.~\ref{fig:archi_dis}, (1), in current methods~\cite{xu2018attngan,li2019controllable,qiao2019mirrorgan}, the discriminator first extracts image features with a small size at $4 \times 4$ through a series of downsampling blocks. Then, these small-size features are used in two ways: one way is to determine whether the image is real or fake, and the other way is to concatenate these image features with the sentence vector to evaluate text-image semantic consistency. So, there are two kinds of loss computed, the unconditional loss and the conditional loss. 

However, concatenating such small-size image features with the coarse sentence vector may not fully evaluate the text-image semantic alignment, because the image features at $4\times 4$ may lose much text-matched image information, and the sentence vector is a global representation of a given text, and cannot comprehensively reflect the alignment between image and text. Based on this, to better evaluate the text-image semantic consistency, we suggest to utilize word embeddings, and combine both text and image information at a shallow layer of the discriminator, instead of at size $4\times 4$. Here, we introduce fusion features, which contain fine-grained text information at word-level and image information at a larger resolution, and then feed this detailed fusion features into the discriminator to evaluate the text-image semantic consistency (see Fig.~\ref{fig:archi_dis}, (2)). To build these fusion features, we convert the output real/fake image features $v \in \mathbb{R}^{C \times H \times W}$ into the same semantic space as word embeddings $w$, to get $v' \in \mathbb{R}^{D \times (H \ast W)}$. Then, the fusion features $F\in \mathbb{R}^{L \times H \times W}$ are obtained by applying a matrix multiplication between $v'$ and $w$, denoted as $F = \text{Reshape}((w)^Tv')$, where each value in $F$ denotes the correlation between each pixel and each word. Finally, we feed fusion features into the discriminator. Basically, checking the difference between fusion features created from word representations and synthetic images and fusion features created from word representations and real images can also reflect the realism of synthetic images, as good fusion features should be realistic and have good text-image semantic alignment, similar to real images matching the corresponding text descriptions. 

\vspace{-2ex}
\subsubsection{One-Way Output.}
\vspace{-1ex}
As mentioned above, there are two kinds of losses computed in the discriminator, unconditional loss for image quality and conditional loss for text-image semantic alignment, see Fig.~\ref{fig:archi_dis} (1). Thus, the discriminator needs to promote the generator to achieve two goals: synthetic images should be indistinguishable from real images, and the semantic alignment between synthetic image and text should be captured similarly as the alignment between real image and text. However, there is no strong connection between these two goals, and thus the output results can be in four kinds of situations (see Fig.~\ref{fig:archi_dis}, (3)): good quality and good semantic alignment, bad quality and good semantic alignment, good quality and bad semantic alignment, bad quality and bad semantic alignment. This is because these two goals can be treated independently, and the network may fail to reach the minimum points of loss function surface for both quality and alignment at the same time. However, thanks to the implementation of fusion features in our discriminator, we can combine these two goals into one. Now, the goal of the discriminator is to promote the generator to produce better fusion features, which should be indistinguishable from the fusion features built from real images and text (see Fig.~\ref{fig:archi_dis}, (4)). Building good fusion features should have a good output image quality and effective correlation between image regions and corresponding finer word information. So, the loss functions are as follows:
\begin{equation}
\footnotesize
\begin{split}
&\mathcal{L}_{G,I}=-{E}_{F'_I\sim P_{g}}\left [ \log(D(F'_{I})) \right ]
\textrm{,} \\
&\mathcal{L}_{D,I}=-{E}\left [\log(D(F_I)) \right ]-{E}_{F'_I\sim P_{g}}\left [ \log(1-D(F'_I)) \right ]
\textrm{,} \\
&\mathcal{L}_{G,V}=-{E}_{F'_V\sim P_{g}}\left [ \log(D(F'_{V})) \right ]
\textrm{,} \\
&\mathcal{L}_{D,V}=-{E}\left [\log(D(F_V)) \right ]-{E}_{F'_V\sim P_{g}}\left [ \log(1-D(F'_V)) \right ]
\textrm{,}
\label{eq:loss}
\end{split}
\end{equation}
where $I$ denotes an image, $V$ denotes a story with a series of images, $F$ are real fusion features created by word representations and real image (or story) features that are sampled from the real distribution, and $F'$ are synthetic fusion features created by word representations and synthetic image (or story) features that are sampled from the model distribution. 

\subsection{Extended Word-Level Spatial Attention}
To generate high-quality story images with finer details and a better consistency, we adopt word-level spatial attention~\cite{xu2018attngan}, and further extend it to focus on all words and visual spatial locations in the entire story. By doing this, our extended attention not only highlights local word-related image regions in the generation process, but also captures the long-range correlation between words in the current sentence and visual spatial locations from other images within the same story, which ensures both the local image region quality and the global consistency of the whole story, tailored for the story visualization task.

First, we convert intermediate visual features $v \in \mathbb{R}^{N \times C \times H \times W}$ into a joint semantic space $\mathbb{R}^{D}$ via a convolution layer, and then reshape the new features to get $v' \in \mathbb{R}^{D \times (H \ast W \ast N)}$, where $N$ is the size of a story, $C$ is the channel number, $H$ is the height of the features, and $W$ is the width. We also reshape word embeddings $w \in \mathbb{R}^{N \times D \times L}$ into $w' \in \mathbb{R}^{D \times (L * N)}$, where $D$ is the feature dimension, and $L$ is the number of words in a sentence. Then, the extended spatial attention $\beta \in \mathbb{R}^{(H \ast W \ast N) \times (L * N)}$ focusing on capturing correlation between words and visual spatial locations across the entire story can be obtained by:
\begin{equation}
\beta_{i,j}=\frac{\text{exp}(a_{i,j})}{\sum^{L*N-1}_{l=0}\text{exp}(a_{i,l})}, \quad\textrm{where}\;
a=v'^{T}w'
\textrm{,}
\end{equation}
where $\beta_{i,j}$ denotes the correlation between the $i$th spatial location and the $j$th word in the story. Therefore, weighted visual features containing word information $v_{w}$ can be obtained by $v_{w} = w'\beta^{T}$. By doing this, the model encourages all images within a story not only to have fine-grained regional details matching corresponding semantic words, but also share similar appearances to make the whole story more consistent.

\section{Experiments}
\label{sec:experiments}
There are a limited number of methods working on the same story visualization task as ours: StoryGAN~\cite{li2019storygan}, CP-CSV~\cite{song2020character}, DUCO~\cite{maharana2021improving}, and VLC~\cite{maharana2021integrating}. StoryGAN is based on generative adversarial networks, and CP-CSV, DUCO, and VLC are built on top of StoryGAN, where CP-CSV relies on character segmentation masks to provide an additional supervision, and DUCO and VLC utilized auxiliary captioning networks to keep consistency. 

\subsection{Datasets}
We adopt Pororo-SV to evaluate our approach, which is first introduced in~\cite{li2019storygan}. Pororo-SV was created from the Pororo dataset~\cite{kim2017deepstory}, which was used for video question answering. There are $13,000$ training pairs and $2,336$ test pairs. Following previous works, we consider every five consecutive images as a story. 

Differently, we do not evaluate our approach on CLEVR-SV~\cite{li2019storygan}, as there are only $15$ different words in the entire dataset, which might fail to fully explore the multi-modal story visualization task. Based on this, we adopt Abstract Scenes~\cite{zitnick2013bringing,zitnick2013learning}. Abstract Scenes was proposed for studying semantic information, which contains over $1,000$ sets of $10$ semantically similar scenes of children playing outside. The scenes are composed with $58$ clip-art objects, and there are six sentences describing different aspects of a scene. 
Following~\cite{zitnick2013learning}, we reserve $1000$ samples as the test set and $497$ samples for validation.

\subsection{Implementation}
Our approach is developed using PyTorch, building on top of the original StoryGAN codebase. The resolution of output images on Pororo-SV is $64 \times 64$, and on Abstract Scenes is $256 \times 256$. The text encoder is a bi-directional LSTM, pretrained to maximize the cosine similarity between matched image and text features~\cite{xu2018attngan}.
We select the best checkpoints and tune hyperparameters by using the FID and FSD scores mentioned below. The network is trained 240 epochs on both Pororo-SV and Abstract Scenes, using the Adam optimizer~\cite{kingma2014adam} with learning rate $0.0002$. All models were trained on a single Quadro RTX 6000 GPU. 

\subsection{Evaluation Metrics}
To evaluate the quality of output images, the Fr\'echet Inception Distance (FID)~\cite{heusel2017gans} is our main evaluation metric, which computes the Fr\'echet distance between the distribution of the synthetic images and ground-truth images in the feature space of a pretrained Inception-v3 network. Besides, following~\cite{song2020character}, we adopt the Fr\'echet Story Distance (FSD), as FID is commonly used to evaluate the image generation task but takes only a single image into account. FSD is a redesigned evaluation matrix for story visualization, and captures more easily the quality of the generated image sequence for a story, with respect to temporal consistency.

However, as both FID and FSD cannot reflect the semantic alignment between sentences and story images, we compute the average cosine similarity (Cosine) between pairs of sentence and synthetic image over the test set, and further scale the value by $100$.

\subsection{Quantitative Evaluation}
Table~\ref{table:quan} shows quantitative comparisons between our approach and the baselines on Pororo-SV and Abstract Scenes. From the table, we can observe that our approach achieves better evaluation results against other methods on both datasets, and builds a new state of the art. Note that differently from CP-CSV, DUCO, and VLC, our approach uses neither segmentation masks nor auxiliary captioning networks. Compared to StoryGAN, our approach achieves 28.7\% improvement in FID, and 44.5\% improvement in FSD on Pororo-SV; and as for Abstract Scenes, our approach achieves 46.5\% improvement in FID, and 73.4\% improvement in FSD. This illustrates that our approach can generate images with finer quality, achieve a better image-text semantic alignment, and keep a higher consistency across story images. Note that we do not evaluate CP-CSV and VLC on Abstract Scenes, as CP-CSV requires segmentation masks, and VLC has not released the complete code.

\begin{table}[t!]
    \centering
    \begingroup
    \setlength{\tabcolsep}{8pt} 
    \caption{Quantitative evaluation between different methods on Pororo-SV and Abstract Scenes. For FID and FSD, lower is better; for text-image cosine similarity (Cosine), higher is better.}
    \label{table:quan}
    \renewcommand{\arraystretch}{1.26} 
    \vspace{-2ex}
    \scalebox{1}{
    \begin{tabular}{l|ccc|ccc}
\hline \hline
    &\multicolumn{3}{c|}{Pororo-SV dataset} &\multicolumn{3}{c}{Abstract dataset}\\
    Method & FID$\downarrow$ & FSD$\downarrow$ & Cosine$\uparrow$ & FID$\downarrow$ & FSD$\downarrow$ & Cosine$\uparrow$ \\
\hline \hline 
    StoryGAN~\cite{li2019storygan} & 78.64 & 94.53 & 0.22 & 135.16 & 55.80 & 3.59 \\
    CP-CSV~\cite{song2020character} & 67.76 & 71.51 & 0.32 & - & - & - \\
    DUCO~\cite{maharana2021improving}  & 95.17 & 171.70 & 0.08 & 142.34 & 49.16 & 3.95\\
    VLC~\cite{maharana2021integrating}  & 94.30 & 122.07 & 0.21 & - & - & - \\
    \hline
    Ours & \textbf{56.08} & \textbf{52.50} & \textbf{2.98} & \textbf{72.34} & \textbf{14.86} & \textbf{4.05} \\
    \hline
    \end{tabular}
     }
    \endgroup
\end{table}

\begin{figure*}[t]
\centering
\begin{minipage}{1\textwidth}
\includegraphics[width=1\linewidth, height=0.615\linewidth]{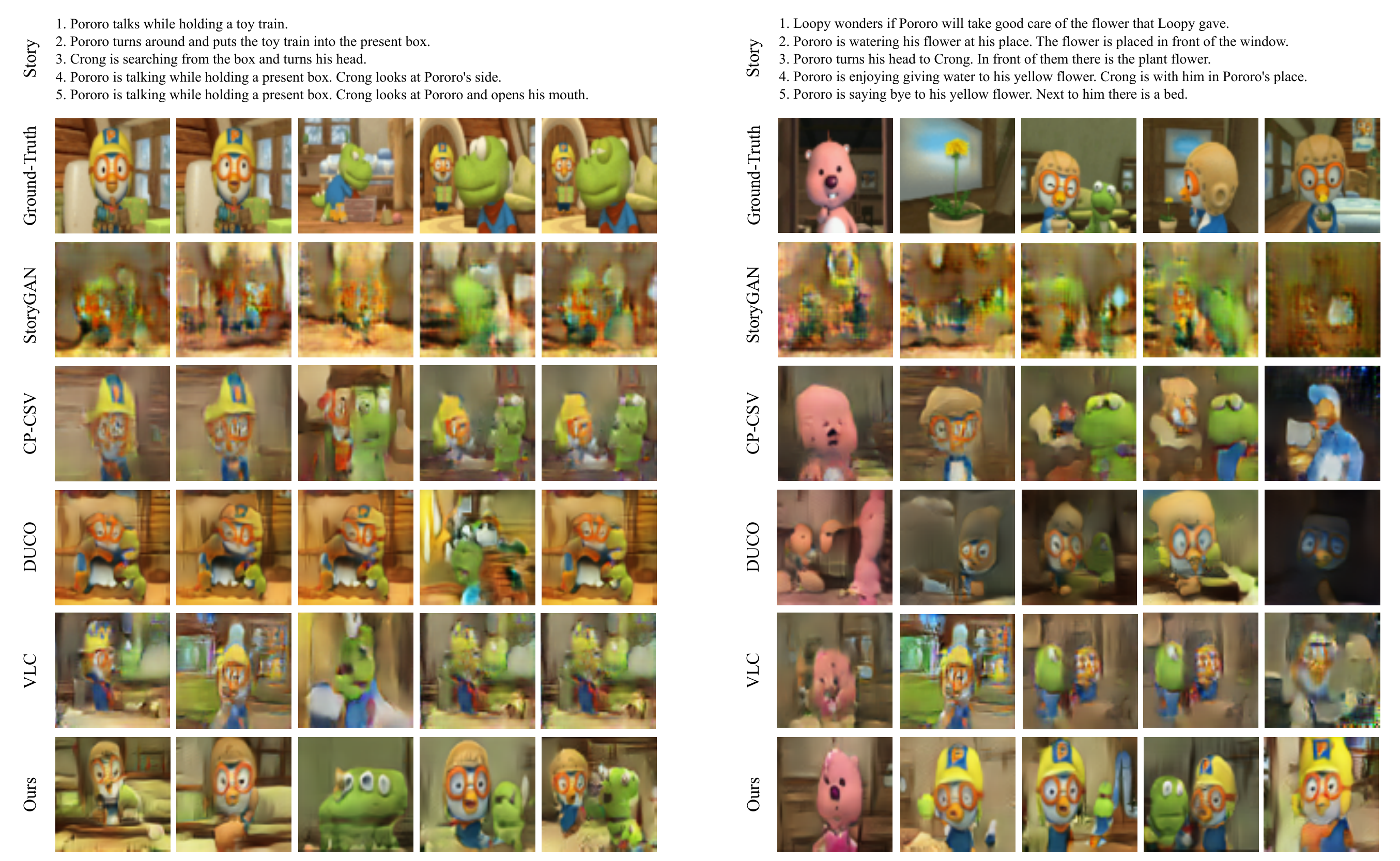}
\end{minipage}

\centering
\vspace{-2ex}
\caption{Comparison between different methods on the Pororo-SV.}
\vspace{-1ex}
\label{fig:qual_pororo}
\end{figure*}

\begin{figure*}[t]
\centering
\begin{minipage}{1\textwidth}
\includegraphics[width=1\linewidth, height=0.474\linewidth]{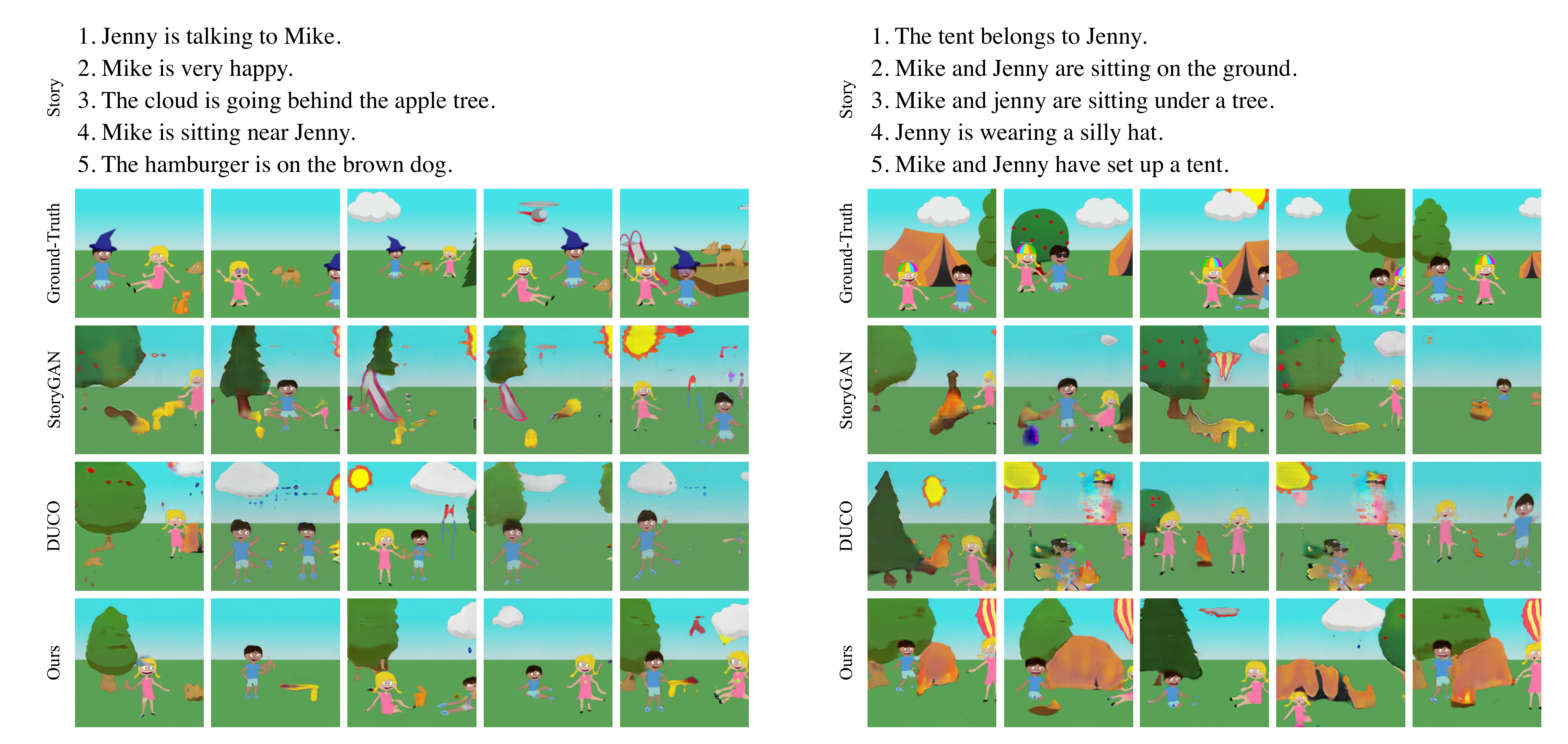}
\end{minipage}

\centering
\vspace{-2ex}
\caption{Comparison between different methods on the Abstract Scenes dataset.}
\label{fig:qual_abstract}
\end{figure*}

\subsection{Qualitative Evaluation}
Figs.~\ref{fig:qual_pororo} and~\ref{fig:qual_abstract} show examples of visual comparisons between our approach and the baselines on Pororo-SV and Abstract Scenes, respectively. As we can observe, on Pororo-SV, our approach generates images with finer regional details, such as detailed character appearances (e.g., penguin Pororo and frog Crong) and sharper shape of objects (e.g., the glasses and the hat on penguin Pororo). For Abstract Scenes, a good improvement can be also observed, where our approach generates high-quality characters (e.g., boy Mike and girl Jenny) with complete and detailed appearances, and also ensures a better consistency, e.g., the word ``tent'' only appears in the first and the last sentences for the second example, but all synthetic images in the story have the object tent, while other methods fails to keep the object tent in the background of all images. 

\subsection{Component Analysis}
\begin{table}[t]
  \centering
  \caption{Component Analysis on Pororo-SV. ``Ours w/o New Sentence Representation'' stands for without using the proposed new sentence representation; ``Ours w/ Discrimiantor'' stands for using the discriminator in current story visualization methods~\cite{li2019storygan,song2020character,maharana2021improving}; ``Ours w/o Extended Spatial Attention'' stands for without adopting the proposed attention; ``Ours w/ Word-Level Spatial Attention'' is with the implementation of word-level spatial attention~\cite{xu2018attngan}, instead of our proposed extended spatial attention.}
  \vspace*{-1ex}
  \label{table:ablation}
  \smallskip
  \scalebox{1}{
  \begin{tabular}{lccc}
    \toprule
    \multicolumn{1}{c}{Method} &
    \multicolumn{1}{c}{\quad\quad FID} & 
    \multicolumn{1}{c}{\quad FSD} & 
    \multicolumn{1}{c}{\quad Cosine} \\
    \midrule
    Ours w/o New Sentence Representation & \quad\quad 68.48 & \quad 62.85 & \quad 2.24 \\
    Ours w/ Discriminator~\cite{li2019storygan} & \quad\quad 62.23 & \quad 59.33 & \quad 2.61 \\
    Ours w/o Extended Spatial Attention & \quad\quad 83.66 & \quad 78.80 & \quad 2.26 \\
    Ours w/ Word-Level Spatial Attention~\cite{xu2018attngan} & \quad\quad 58.26 & \quad 63.39 & \quad 2.54 \\
    \midrule
    Ours w/ Pretrained BERT & \quad\quad 52.38 & \quad49.69 & \quad 3.71\\
    Ours w/ FT BERT & \quad\quad 50.96 & \quad 48.81 & \quad 3.95\\
    Ours w/ BERT Scratch & \quad\quad 55.78 & \quad 51.71 & \quad 2.80 \\
    \midrule
    Ours & \quad\quad 56.08 & \quad 52.50 & \quad 2.98 \\
    \bottomrule
  \end{tabular}
  }
\end{table}
We conduct an ablation study to evaluate the effectiveness of different components proposed in the paper, and the results are shown in Table~\ref{table:ablation}. 

\subsubsection{Sentence Representation with Word Information.} First, we observe that the new sentence representation improves scores on all evaluation metrics. We attribute this improvement to the better representation of text that contains fine-grained word information from the entire story, which enables high-quality initial image features with finer regional details and a better consistency (e.g., better FSD score), and thus further improves final synthetic results (e.g., better FID and Cosine scores). This can be also supported by the observation found in text-to-image generation~\cite{zhu2019dm}, where the quality of initial image features can considerably affect the quality of output images in such a sequential upsampling generation pipeline (see Fig.~\ref{fig:archi}).

\subsubsection{Discriminator with Fusion Features and Extended Word-Level Spatial Attention.} A degradation can be observed when our approach does not implement the proposed discriminator. Similarly, without adopting proposed extended spatial attention, we can observe a decrease in both FID and FSD scores. We think that this is because both components complement each other: (1)~without the proposed discriminator, although the generator tries to make use of word-level information via the proposed attention in the generation process, there is no corresponding word-level feedback from the discriminator, regarding whether word-related attributes are produced in synthetic story images, and (2) without the extended spatial attention, the fine-grained feedback from the proposed discriminator at both word- and image region-level cannot be fully utilized by the generator. 
Therefore, both components can work together to make a full use of word information to enable a higher-quality image generation with a better story consistency. Besides, we observe that the scores for ``Ours w/o New Sentence Representation'' are better than for ``Ours w/o Extended Spatial Attention''. This shows that the fine-grained training feedback provided by our discriminator can be more utilized by the proposed extended word-level spatial attention.

Besides, we further record the number of trainable parameters in both kinds of discriminators, where our proposed single-directional discriminator only contains 10.4M trainable parameters, while the number of parameters in StoryGAN, DUCO, and VLC is 47.2M, and 70.9M for CP-CSV. So, compared to StoryGAN, our proposed discriminator reduces about 77.97\% number of parameters, but achieves 28.7\% improvement in FID, and 44.5\% improvement in FSD on Pororo-SV, which verifies its superiority.

Moreover, we replace the extended spatial attention with word-level spatial attention~\cite{xu2018attngan} to verify its effectiveness. The main improvement using the extended spatial attention is shown on the consistency of output results, especially on FSD score. This is because our proposed component focuses mainly on the whole story with multiple images and sentences, instead of only a pair of image and sentence, which coincides with the purpose of keeping the consistency across story images.

\subsubsection{Text Encoder Using BERT.} 
In Table~\ref{table:ablation}, we use BERT to replace LSTM as the text encoder. If the BERT is trained from scratch, denoted as ``w/ BERT Scratch'', there is no significant changes on all metrics. We think that this is because Pororo-SV is small and each caption contains a limited number of words, and thus an RNN-based text encoder (e.g., LSTM) is strong enough to capture long-range dependencies between words.
If we use a pretrained BERT, denoted as ``w/ Pretrained BERT'' or further fine-tune BERT on Pororo-SV, denoted as ``w/ FT BERT'', performance improves. This is because BERT pretrained on a larger text corpus (e.g., Wikipedia) can generate better text representations.

\subsection{Visualization of Extended Spatial Attention and New Sentence Representation}
\begin{figure}[t]
\begin{minipage}{1\textwidth}
\includegraphics[width=1\linewidth, height=0.5\linewidth]{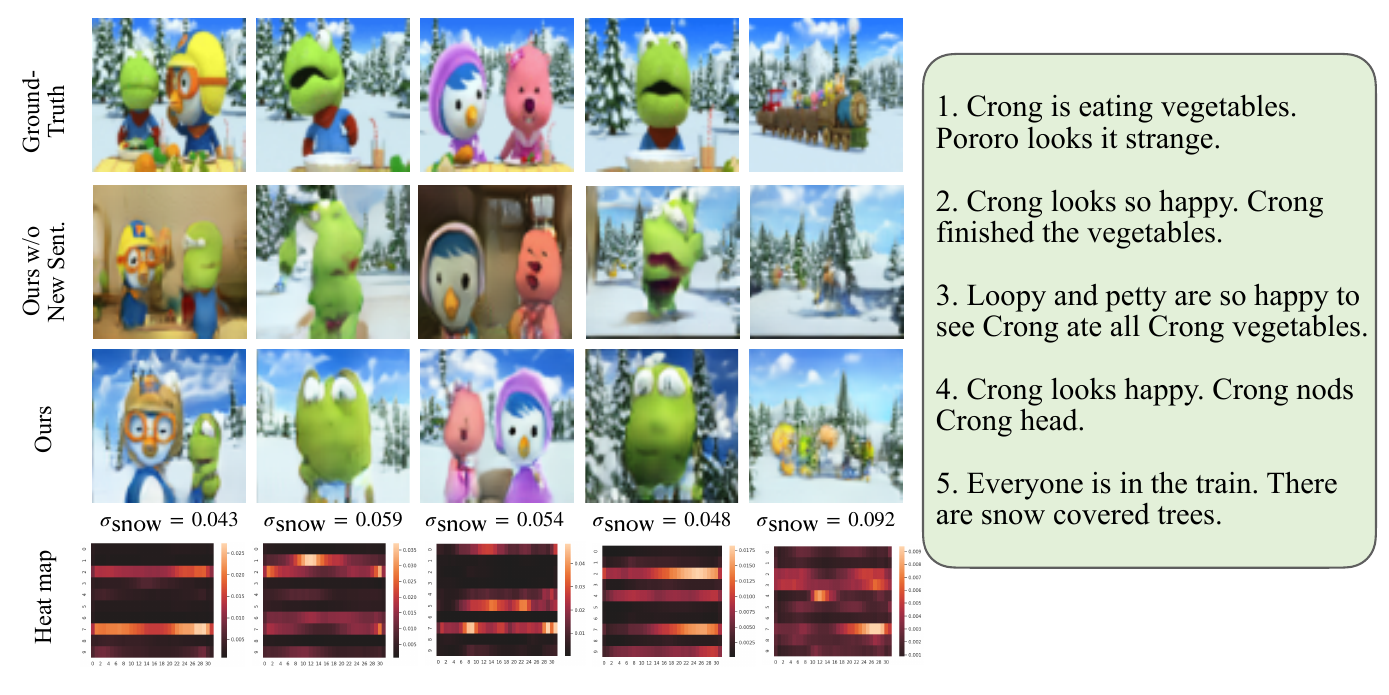}
\end{minipage}

\centering
\vspace{-2ex}
\caption{Visualization of extended spatial attention and new sentence representation.}
\label{fig:heatmap}
\end{figure}
Capturing relations between words and image regions has been adopted in text-to-image generation~\cite{xu2018attngan,li2019controllable,qiao2019mirrorgan} and image editing using text~\cite{li2020manigan,li2020lightweight} to improve image quality and text-image semantic alignment. Our proposed extended word-level spatial attention captures relations between image regions and words from the entire story to ensure local image quality and also global consistency across story images, tailored for story visualization. In Fig.~\ref{fig:heatmap}, we visualize the extended spatial attention using a heatmap. The $y$ axis denotes the last 10 words in the story (i.e., sentence 5: Everyone\ldots), and $x$ denotes image regions, where we evenly split an image into 32 regions and compute the average for each region. So, the heatmap shows the correlation values captured by our attention between the last 10 words and image regions in each image. The word {snow} ($y=7$) is highlighted in each heatmap, and {snow} is also generated in all output images, which verifies that our method captures distant word-image relations and ensures consistency even when some keywords do not appear in all sentences.

Furthermore, we also visualize the proposed new sentence representation with word information by showing the $\sigma$ values between ``snow'' and each sentence in Fig.~\ref{fig:heatmap}. As we can see, our sentence representation learns to give high $\sigma$ to keywords, without it, distant images from keywords may not keep consistency (e.g., there is no snow in the $1$st image of ``Ours w/o New Sent.''). 

\subsection{General Applicability of New Sentence Representation}
\begin{table}[t]
  \centering
  \caption{Effects of our new sentence representation adopted in story visualization on Pororo-SV and text-to-image generation on CUB birds. For FID and FSD, lower is better, for IS, higher is better. ``+ New Sent.'' means using our new representation.}
  \vspace*{-1ex}
  \label{table:sent}
  \smallskip
  \scalebox{1}{
  \begin{tabular}{lccc}
    \toprule
    \multicolumn{1}{c}{Method} &
    \multicolumn{1}{c}{\quad\quad FID} & 
    \multicolumn{1}{c}{\quad FSD} & 
    \multicolumn{1}{c}{\quad IS} \\
    \midrule
    StoryGAN + New Sent. & \quad\quad 72.81 & \quad 84.06 & \quad - \\
    CP-CSV + New Sent. & \quad\quad 63.12 & \quad 64.29 & \quad - \\
    DUCO + New Sent. & \quad\quad 87.82 & \quad 131.83 & \quad - \\
    VLC + New Sent. & \quad\quad 82.19 & \quad 100.94 & \quad - \\
    \midrule
    AttnGAN~\cite{xu2018attngan} & \quad\quad 23.98 & \quad - & \quad 4.36 \\
    AttnGAN + New Sent. & \quad\quad 19.20 & \quad - & \quad 4.71 \\
    DFGAN~\cite{tao2020df} & \quad\quad 14.81 &  \quad - & \quad 5.10 \\
    DFGAN + New Sent. & \quad\quad 11.98 & \quad - & \quad 5.16\\
    \midrule
    Ours & \quad\quad 56.08 & \quad 52.50 & \quad - \\
    \bottomrule
  \end{tabular}
  }
\end{table}

As shown in Table~\ref{table:sent}, we implement our new sentence representation in other story visualization methods and also text-to-image generation (e.g., AttnGAN~\cite{xu2018attngan} and DFGAN~\cite{tao2020df}) by feeding word information into the sentence vector without changing networks. We can observe that our sentence representation further improves these methods by having better evaluation scores. For story visualization, the improvement is mainly because our proposed sentence representation enables high-quality initial image features with finer regional details and a better consistency, and thus further improves final synthetic results. For text-to-image generation, the improvement verifies the observation shown in the work~\cite{zhu2019dm} that the quality of initial image features can considerably affect the quality of output images in a sequential upsampling generation pipeline.

\subsection{Human Evaluation}
\begin{table}[t]
  \centering
  \caption{Results of a side-by-side human evaluation between our approach and DUCO on Pororo-SV.}
  \vspace*{-1ex}
  \label{table:human_comp}
  \smallskip
  \scalebox{1}{
  \begin{tabular}{lccc}
    \toprule
    \multicolumn{1}{c}{Choice (\%)} &
    \multicolumn{1}{c}{\quad\quad Ours} & 
    \multicolumn{1}{c}{\quad Tie} & 
    \multicolumn{1}{c}{\quad DUCO} \\
    \midrule
    Visual Quality & \quad\quad 75.3  &\quad 8.3 &\quad 16.4    \\
    \midrule
    Consistency &\quad\quad 81.8 &\quad 4.4 &\quad 13.8  \\
    \midrule
    Relevant  &\quad\quad 72.3 &\quad 5.7 &\quad 22.0  \\
    \bottomrule
  \end{tabular}
  }
\end{table}
We also conduct a human evaluation to comprehensively evaluate the performance of our approach on Pororo-SV. To do this, following~\cite{li2019storygan}, a side-by-side human evaluation study is conducted based on three evaluation criteria: (1) the visual image quality, (2) the consistency among story images, and (3) the semantic alignment between paired image and sentence.
The study compares the synthetic image sequences from our approach and DUCO. We showed the generated story images along with the corresponding story sentences from two methods, and three options are provided: (1) the first is better, (2) two are similar, and (3)~the second is better.
In this evaluation, we randomly choose 200 synthetic story images, generated from corresponding story descriptions sampled from the testing dataset, and each was assigned to 5 workers to reduce human variance. 

From Table~\ref{table:human_comp}, story images produced by our approach are more preferred by workers on three evaluate criteria, which demonstrates the effectiveness of our approach on high-quality image generation with a better text-image semantic alignment, and a better story consistency.

\section{Conclusion}
In this paper, we investigated the task of story visualization. We proposed a new approach by exploring fine-grained information at both word and story level, and demonstrate the effectiveness of proposed components by conducting extensive experiments. Our approach has three novel components: a new sentence representation selectively incorporates different word information from the entire story to enable a better consistency, the extended spatial attention captures relations between image regions and words across the entire story to enable the generation of higher-quality story images with better consistency, and the novel discriminator contains a smaller number of parameters, but effectively provides the generator with fine-grained training signals, evaluating the quality of fusion features to examine whether word-related image attributes are generated in story images and also their quality, and thus encouraging the generator with the proposed attention to highlight image relations corresponding to semantic words from the story. Finally, a human evaluation has been conducted to further verify the effective performance of our proposed approach.

As for future work, there is still space for quality improvement on different datasets, such as the images details on characters. Also, a more concise architecture design is desired, as current methods (including our approach) require a large amount of memory storage to hold and train. Also, the resolution of output images is small at $64 \times 64$, which is less practical for real-word applications. 

\section*{Acknowlegments}
This work was supported by the Alan Turing Institute under the EPSRC grant EP/N510129/1, by the AXA Research Fund, and by the EPSRC grant EP/R013667/1. We also acknowledge the use of the EPSRC-funded Tier 2 facility JADE (EP/P020275/1) and GPU computing support by Scan Computers International Ltd.

%
%
\bibliographystyle{splncs04}
\bibliography{egbib}
\end{document}